%% file: camera_ready_paper.tex
\documentclass{article}
\usepackage{amsmath,amssymb,amsfonts}
\usepackage{mathtools}
\usepackage{color}
\usepackage{array}
\usepackage{stfloats}
\usepackage{nccmath}
\usepackage{setspace}
\usepackage{mwe}
\usepackage{textcomp}
\usepackage{xcolor}
\usepackage{microtype}
\usepackage{xspace}
\usepackage{graphicx}
\usepackage{subfigure}
\usepackage{booktabs}
\usepackage{hyperref}

\usepackage[accepted]{icml2019}

\icmltitlerunning{Adversarial Imitation Learning via Random Search in Lane Change Decision-Making}
\DeclareUnicodeCharacter{2212}{-}
\begin{document}

\twocolumn[
\icmltitle{Adversarial Imitation Learning via Random Search in \\ Lane Change Decision-Making}




\begin{icmlauthorlist}
\icmlauthor{MyungJae Shin}{to}
\icmlauthor{Joongheon Kim}{to}
\end{icmlauthorlist}

\icmlaffiliation{to}{School of Computer Science and Engineering, Chung-Ang University, Seoul, Republic of Korea}

\icmlcorrespondingauthor{MyungJae Shin}{mjshin.cau@gmail.com}
\icmlcorrespondingauthor{Joongheon Kim}{joongheon@gmail.com}

\icmlkeywords{Machine Learning, ICML}

\vskip 0.3in
]



\printAffiliationsAndNotice{}  % leave blank if no need to mention equal contribution

\begin{abstract}
As the advanced driver assistance system (ADAS) functions become more sophisticated, the strategies that properly coordinate interaction and communication among the ADAS functions are required for autonomous driving. This paper proposes a derivative-free optimization based imitation learning method for the decision maker that coordinates the proper ADAS functions. The proposed method is able to make decisions in multi-lane highways timely with the LIDAR data. The simulation-based evaluation verifies that the proposed method presents desired performance. Note that this framework is accepted to be published in proceedings of IJCNN 2019 and IJCAI 2019.
\end{abstract}

\input{1Introduction.tex}

\input{2method.tex}
\input{3experiments.tex}
%\input{(4)conclusion.tex}

\section*{Acknowledgements}
This research was supported by IITP grant funded by the Korea government (MSIP) (No. 2017-0-00068, A Development of Driving Decision Engine for Autonomous Driving using Driving Experience Information).  Note that the full versions of this paper are appeared in proceedings of IJCNN 2019~\cite{shin2019imitate} and IJCAI~\cite{shin2019rail}.


\nocite{langley00}
\nocite{shin2019imitate}
\bibliography{5reference}
\bibliographystyle{icml2019}

\end{document}

%% file: 1Introduction.tex
\section{Introduction}\label{sec1}
\begin{figure}[t!]
    \centering
    \includegraphics[width=0.99\columnwidth]{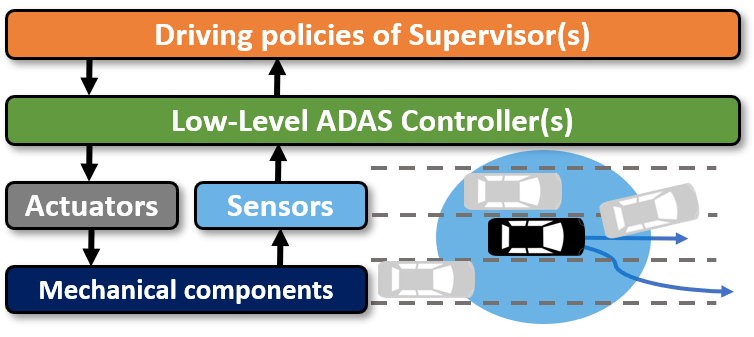}
    \caption{Hierarchical control for highly interdependent autonomous vehicle systems. }
    \label{fig:overview}
    \vspace{-6mm}
\end{figure}

% Autonomous Driving 에 관한 서술
Ever since the introduction of autonomous vehicles, autonomy of vehicles has been a subject of great interest among researchers. Deep reinforcement learning (DRL) has been considered as one of feasible solutions to replace human involvement with autonomous control systems. 
As the advanced driver assistance system (ADAS) functions become more complex, the strategies that properly coordinate interaction and communication among the ADAS functions are required for autonomous driving~\cite{cioran2015system}. 
The architecture of ADAS enabled autonomous vehicle control is in Fig.~\ref{fig:overview}. The low-level ADAS controllers are connected to the sensors (i.e., LIDAR sensor) directly. The controllers determine the data from sensors to grasp the situation and transmit the determined operations to mechanical components through actuators. The systems of the autonomous vehicles are managed by a supervisor that coordinates the low-level controllers. The objective of the supervisor is to be a decision maker for the overall system that consists of ADAS functions during driving. Especially, lane change decision-making is one of the challenging problems in this research. It is essential to form efficient long term driving assistance strategies in limited situations such as multi-lane highway environments.
% 강화학습에 관한 서술
Recent work focuses on DRL to make the driving policies of supervisor to be lane change decision-maker in highway scenarios~\cite{hoel2018automated}. However, when the driving policies are trained through DRL, the problem of safety as well as the robustness of the trained policy is caused by the presence of undesirable policy to maximize the expected rewards at the expense of violating the implicit rules of the environments~\cite{pan2018agile}. These problems motivate the researchers to adopt imitation learning (IL) to optimize the driving policy. Among IL frameworks, generative adversarial imitation learning (GAIL) has shown remarkable performance in the areas of robotics, autonomous vehicles, and etc~\cite{pomerleau1991rapidly}. 
However, since GAIL framework uses trust region policy optimization (TRPO), a large amount of data is required to achieve remarkable performance~\cite{schulman2015trust}. To address this issue, the DRL algorithms that optimize policy in GAIL framework become complicated; and thus the models ultimately lead to reproducibility crisis. Recently, augmented random search (ARS) that consists of the natural gradient policy algorithm is proposed~\cite{ars}. In this work, we present an IL-based method \textit{adversarial imitation learning via random search (AILRS)} that combines the concepts of ARS and GAIL. Since AILRS is based on derivative-free policy optimization, it is relatively easy to reconfigure the robust trained policy~\cite{shin2019imitate,shin2019rail}.

%% file: 2method.tex
\section{AILRS}\label{sec2}
\begin{algorithm}[t!]
   \caption{Adversarial imitation learning via random search (AILRS)}
   \label{alg:AILRS}
\begin{algorithmic}
\footnotesize
   \STATE {\bfseries Input:} step size $\alpha$, number of sampled directions $N$, $K$ number of directions to update
   \STATE {\bfseries Initialize:} $\theta_0=\mathbf{0}\in\mathbb{R}^{p \times n}$,$\mu_0 = \mathbf{0}\in \mathbb{R}^n$; $\sum_{0} = \mathbf{I}_n \in \mathbb{R}^{n\times n}$
   \REPEAT
    \STATE Sample $\bold{\delta_t} = \left\{ \delta_1, \delta_2, . . . , \delta_N ; \delta_i \in \mathbb{R}^{p \times n} \right\} $ with i.i.d.
    \STATE Collect $2N$ rollouts and their corresponding rewards using the $2N$ policies.
    \STATE $\pi_{t,+\delta_{i}}(s)  = (\theta_t + \nu\delta_i) diag (\sum_t)^{−1/2}(s - \mu_t )$
    \STATE $\pi_{t,-\delta_{i}}(s)  = (\theta_t - \nu\delta_i) diag (\sum_t)^{−1/2}(s - \mu_t )$ 
    \STATE for $i \in  \left\{ 1, 2, \dots, N \right\}$
    \STATE {\bfseries Update discriminator parameter $\phi_t$ :}
         
    \STATE $\nabla_{\phi_t}L_{LS} = \frac{1}{2}\mathbb{E}_{\pi_E} \left[(\nabla_{\phi_t}\mathcal{D}_{\phi_{t}}(s, a) - b)^2\right]$
    \STATE $+\frac{1}{2}\mathbb{E}_{\pi_\theta}\left[(\nabla_{\phi}\mathcal{D}_{\phi_{t}}(s,a) - a)^2\right]$
         
    \STATE {\bfseries Update the policy parameter $\theta_t$ :}
    \STATE $\theta_{t+1} = \theta_{t} + \frac{\alpha}{N\sigma_R}\sum^{K}_{i=1}{\left[ r(\pi_{t,+\delta_{i}}(s) ) - r(\pi_{t,-\delta_{i}}(s)) \right]\delta_{(i)} }$ 
    \STATE where trajectories $T$ sampled from $\pi_{t,\pm\delta_{i}}$ 
    \STATE and $r(\pi_{t,(i),\pm}) = \mathbb{E}_{(s,a)\thicksim \pi_{t,\pm\delta_{i}}} [ -\log( 1 - \mathcal{D}_{\phi_{t}}(T))]$
    
    \STATE Set $\mu_{t+1}$, $\sum_{t+1}$ to be the mean and covariance of the states encountered from the start of training
    \STATE $t$ = $t + 1$
   \UNTIL{$t\le$ Max Iteration}
\end{algorithmic}
\end{algorithm}
\vspace{-2mm}

In this section, the proposed method, called \textit{Adversarial imitation learning via random search (AILRS)}, is briefly introduced~\cite{shin2019imitate,shin2019rail}. As shown in Algorithm~\ref{alg:AILRS}, the finite differences are used to adjust a parameterized linear policy. To train the weights of policy $\pi_\theta$, a random matrix with a small value is added to $\theta$. The matrix with the same value is subtracted to $\theta$. As a result, two temporal weights $\pi_{t,+\delta_{i}}(s)$ and $\pi_{t,i,-}(s)$ are generated; the trajectories of state-action pairs are collected through these weights. The discriminator returns the probability of classifying the trajectories from expert; and it is used as rewards. 

%% file: 3experiments.tex
\vspace{-2mm}
\section{Experiments}\label{sec3}
\begin{figure}[t!]
    \centering
    \includegraphics[width=0.90\columnwidth]{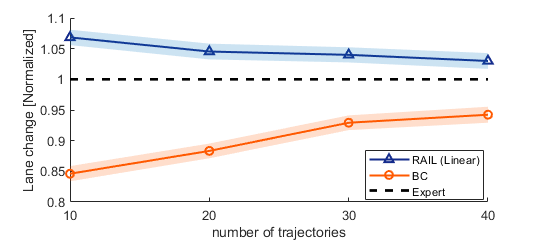}
    \caption{Normalized number of lane changes.}
    \label{fig:lane}
\vspace{-3mm}
\end{figure}

\begin{figure}[t!]
    \centering
    \includegraphics[width=0.90\columnwidth]{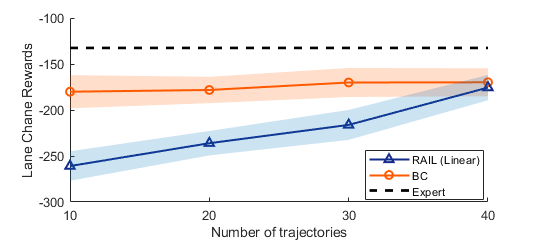}
    \caption{Rewards of lane changes.}
    \label{fig:laner}
\vspace{-3mm}
\end{figure}
The host vehicle continuously obtains lane change rewards during driving. The number of lane change in Fig.~\ref{fig:lane} has a different tendency from the lane change reward in Fig.~\ref{fig:laner}. The lane change cannot be done in a single determined action. Until the lane change is completed, the host vehicle can change the decision according to the observation. If more than half of the host vehicles do not cross the lane, the lane change reward increases whereas the number of lane changes does not. Therefore, 
the trained policy through AILRS shows more number of lane changes whereas it presents the smaller lane change reward, comparing to expert. This is because the trained policy changes more lanes than expert's behaviors. However, BC has a small number of lane changes as well as a small lane change reward due to frequent decision changes.  

%% file: camera_ready_paper.bbl
\begin{thebibliography}{8}
\providecommand{\natexlab}[1]{#1}
\providecommand{\url}[1]{\texttt{#1}}
\expandafter\ifx\csname urlstyle\endcsname\relax
  \providecommand{\doi}[1]{doi: #1}\else
  \providecommand{\doi}{doi: \begingroup \urlstyle{rm}\Url}\fi

\bibitem[Cioran(2015)]{cioran2015system}
Cioran, A.
\newblock System {I}ntegration {T}esting of {A}dvanced {D}river {A}ssistance
  {S}ystems, 2015.

\bibitem[Hoel et~al.(2018)Hoel, Wolff, and Laine]{hoel2018automated}
Hoel, C.-J., Wolff, K., and Laine, L.
\newblock Automated {S}peed and {L}ane {C}hange {D}ecision {M}aking using
  {D}eep {R}einforcement {L}earning.
\newblock \emph{arXiv:1803.10056}, 2018.

\bibitem[Mania et~al.(2018)Mania, Guy, and Recht]{ars}
Mania, H., Guy, A., and Recht, B.
\newblock Simple random search provides a competitive approach to reinforcement
  learning.
\newblock \emph{arXiv}, 2018.

\bibitem[Pan et~al.(2018)Pan, Cheng, Saigol, Lee, Yan, Theodorou, and
  Boots]{pan2018agile}
Pan, Y., Cheng, C.-A., Saigol, K., Lee, K., Yan, X., Theodorou, E., and Boots,
  B.
\newblock Agile autonomous driving using end-to-end deep imitation learning.
\newblock \emph{RSS}, 2018.

\bibitem[Pomerleau(1991)]{pomerleau1991rapidly}
Pomerleau, D.
\newblock Rapidly adapting artificial neural networks for autonomous
  navigation.
\newblock In \emph{NIPS}, pp.\  429--435, 1991.

\bibitem[Schulman et~al.(2015)Schulman, Levine, Abbeel, Jordan, and
  Moritz]{schulman2015trust}
Schulman, J., Levine, S., Abbeel, P., Jordan, M., and Moritz, P.
\newblock Trust region policy optimization.
\newblock In \emph{ICML}, 2015.

\bibitem[Shin \& Kim(2019{\natexlab{a}})Shin and Kim]{shin2019imitate}
Shin, M. and Kim, J.
\newblock Adversarial {I}mitation {L}earning via {R}andom {S}earch.
\newblock In \emph{IJCNN}, 2019{\natexlab{a}}.

\bibitem[Shin \& Kim(2019{\natexlab{b}})Shin and Kim]{shin2019rail}
Shin, M. and Kim, J.
\newblock Randomized {A}dversarial {I}mitation {L}earning for {A}utonomous
  {D}riving.
\newblock In \emph{IJCAI}, 2019{\natexlab{b}}.

\end{thebibliography}
